\documentclass[runningheads]{llncs}
\pdfoutput=1
\usepackage{graphicx}

\usepackage{tikz}
\usepackage{comment}
\usepackage{amsmath,amssymb} 
\usepackage{color}
\usepackage{array}
\usepackage{tabularx}

\usepackage[accsupp]{axessibility}  

\begin{document}
\pagestyle{headings}
\mainmatter
\def\ECCVSubNumber{*}  

\title{NeuRIS: Neural Reconstruction of Indoor Scenes Using Normal Priors} 

\titlerunning{NeuRIS}
%
\author{Jiepeng Wang\inst{1} \and
Peng Wang\inst{1} \and
Xiaoxiao Long\inst{1} \and
Christian Theobalt \inst{2} \and
Taku Komura \inst{1} \and
Lingjie Liu$^{\ast}$ \inst{2} \and
Wenping Wang\thanks{Corresponding authors.}\inst{3}
}
\authorrunning{J. Wang et al.}
%
\institute{
The University of Hong Kong \email{\{jpwang,pwang3,xxlong,taku\}@cs.hku.hk} \and
Max Planck Institute for Informatics  \email{\{lliu,theobalt\}@mpi-inf.mpg.de}
\and
Texas A\&M University \email{wenping@tamu.edu}
}

\maketitle



\definecolor{purple}{cmyk}{0.45,0.86,0,0}
\definecolor{bleudefrance}{rgb}{0.19, 0.55, 0.91}
\definecolor{darkorange}{rgb}{1, 0.55, 0}
\definecolor{limegreen}{rgb}{0.2, 0.8, 0.2}

\def\etal{et al.}			  
\def\eg{e.g.,~}               
\def\ie{i.e.,~}               
\def\etc{etc}                 
\def\cf{cf.~}                 
\def\viz{viz.~}               
\def\vs{vs.~}                 


\newlength\paramargin
\newlength\figmargin
\newlength\secmargin
\newlength\figcapmargin

\setlength{\secmargin}{0.0mm}
\setlength{\paramargin}{0.0mm}
\setlength{\figmargin}{0.0mm}
\setlength{\figcapmargin}{0.5mm}

\newcommand{\red}{\textcolor{red}}
\newcommand{\blue}{\textcolor{blue}}

\newcommand{\mpage}[2]
{
\begin{minipage}{#1\linewidth}\centering
#2
\end{minipage}
}

\newcommand{\mfigure}[2]
{
\begin{subfigure}[b]{#1\linewidth}\centering
\includegraphics[width=\linewidth]{#2}
\end{subfigure}
}

\newcommand{\Paragraph}[1]
{
\vspace{\paramargin}
\paragraph{#1}
}

\newcommand{\heading}[1]
{
\vspace{1mm}
\noindent \textbf{#1}
}   

\newcommand{\secref}[1]{Section~\ref{#1}}
\newcommand{\figref}[1]{Figure~\ref{#1}} 
\newcommand{\tblref}[1]{Table~\ref{#1}}
\newcommand{\eqnref}[1]{Equation~\ref{#1}}
\newcommand{\thmref}[1]{Theorem~\ref{#1}}
\newcommand{\prgref}[1]{Program~\ref{#1}}
\newcommand{\algref}[1]{Algorithm~\ref{#1}}
\newcommand{\clmref}[1]{Claim~\ref{#1}}
\newcommand{\lemref}[1]{Lemma~\ref{#1}}
\newcommand{\ptyref}[1]{Property~\ref{#1}}

\long\def\ignorethis#1{}
\newcommand {\todo}{{\textbf{\color{red}[TO-DO]\_}}}
\def\newtext#1{\textcolor{blue}{#1}}
\def\modtext#1{\textcolor{red}{#1}}

\newcommand{\cy}[1]{{\color{purple}          {#1}}}
\newcommand{\cyc}[1]{{\color{purple}          {[CY: #1]}}}
\newcommand{\wpg}[1]{{\color{orange}          {#1}}}
\newcommand{\wpeng}[1]{{\color{orange}          {[Peng: #1]}}}
\newcommand{\xlc}[1]{{\color{cyan}          {[Xin: #1]}}}
\newcommand{\xl}[1]{{\color{cyan}          {#1}}}
\newcommand{\LXX}[1]{{\color{cyan}          {[Xiao: #1]}}}
\newcommand{\ctc}[1]{{\color{darkorange}  {[CT: #1]}}}
\newcommand{\JP}[1]{{\color{blue}  {[JP: #1]}}}
\newcommand{\LJ}[1]{{\color{red}  {[LJ: #1]}}}
\newcommand{\JPW}[1]{{\color{blue}  {#1}}}
\newcommand{\NOTE}[1]{{\color{red}  {#1}}}
\newcommand{\maincontent}[1]{\textbf{\textcolor{blue}{#1}}}
\newcommand{\rem}[1]{{\color{red}  {#1}}}
\newcommand{\tb}[1]{\textbf{#1}}
\newcommand{\mb}[1]{\mathbf{#1}}
\newcommand{\revised}[1]{\textcolor{red}{#1}}

\newcommand{\jbox}[2]{
  \fbox{%
  	\begin{minipage}{#1}%
  		\hfill\vspace{#2}%
  	\end{minipage}%
  }}

\newcommand{\jblock}[2]{%
	\begin{minipage}[t]{#1}\vspace{0cm}\centering%
	#2%
	\end{minipage}%
}
\begin{abstract}
Reconstructing 3D indoor scenes from 2D images is an important task in many computer vision and graphics applications. A main challenge in this task is that large texture-less areas in typical indoor scenes make existing methods struggle to produce satisfactory reconstruction results. We propose a new method, named {\em NeuRIS}, for high-quality reconstruction of indoor scenes. 
The key idea of {\em NeuRIS} is to integrate estimated normal of indoor scenes as a prior in a neural rendering framework for reconstructing large texture-less shapes and, importantly, to do this in an adaptive manner to also enable the reconstruction of irregular shapes with fine details. 
Specifically, we evaluate the faithfulness of the normal priors on-the-fly by checking the multi-view consistency of reconstruction during the optimization process. 
Only the normal priors accepted as faithful will be utilized for 3D reconstruction, which typically happens in the regions of smooth shapes possibly with weak texture.
However, for those regions with small objects or thin structures, for which the normal priors are usually unreliable, we will only rely on visual features of the input images, since such regions typically contain relatively rich visual features (e.g., shade changes and boundary contours).
Extensive experiments show that {\em NeuRIS} significantly outperforms the state-of-the-art methods in terms of reconstruction quality\footnote{Our project page: https://jiepengwang.github.io/NeuRIS/}.


\end{abstract}
\section{Introduction}

Reconstructing 3D indoor scenes from multiple input images is an important and challenging task in many practical applications, such as robotic navigation, virtual reality and path planning. Indoor scenes usually contain many large texture-less areas and repetitive patterns, such as white walls, floors, and reflecting surfaces, which is challenging for applying conventional matching-based dense reconstruction algorithms~\cite{schoenberger2016colmap,zheng2014patchmatch,shen2013openmvs} that heavily rely on the correspondence of distinct visual features, leading to poor reconstruction results.\begin{figure}
  \centering
  \includegraphics[width=\textwidth]{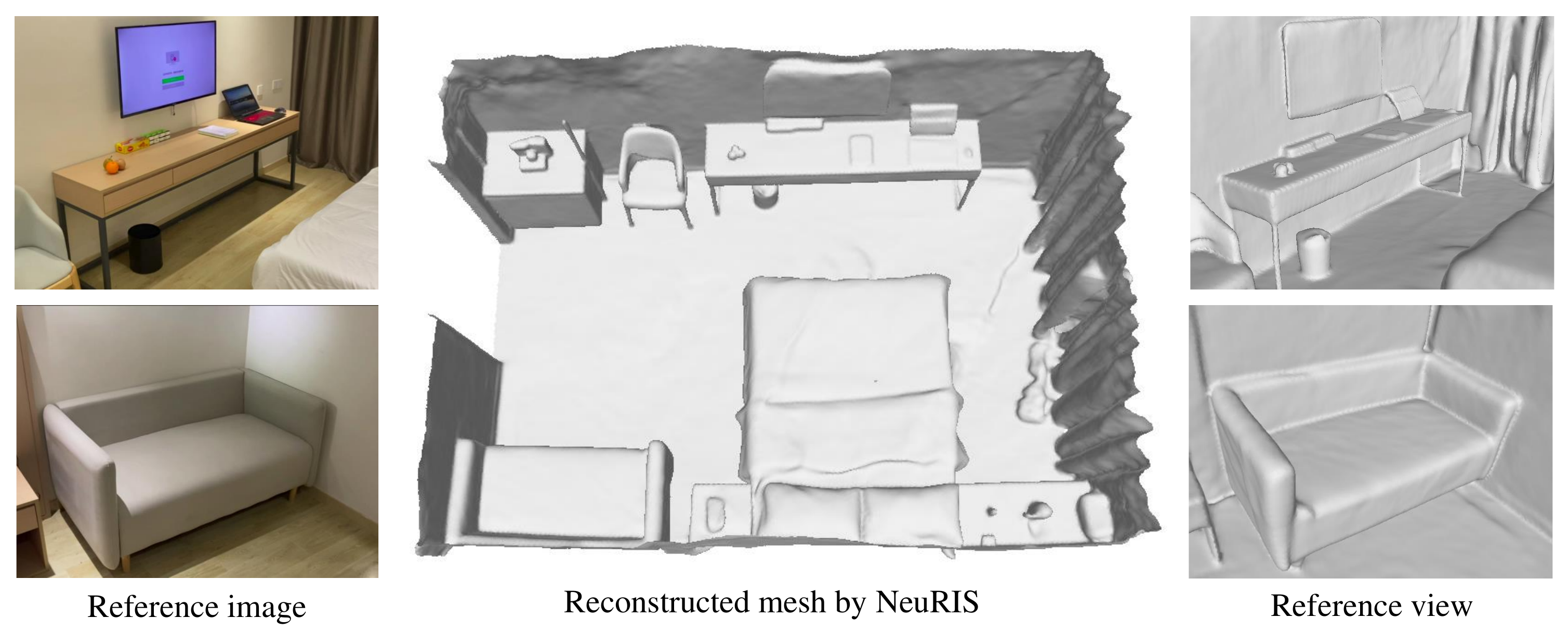}
  \caption{\textbf{Room-scale indoor reconstruction.} Given a set of images equally sampled from a video (captured by iPhone 11), NeuRIS succeeds in reconstructing smooth surfaces with fine details. Note the large-area flat regions (like the floor, the sofa) and the delicate structures (like the chair legs and the desk) in our reconstructed mesh.}
  \label{fig:teaser}
\end{figure}With the success of deep neural networks, data-driven (depth-based and TSDF-based) methods~\cite{shen2018mvdepthnet,long2020cnmnet,im2019dpsnet,teed2018deepv2d,murez2020atlas,sun2021neuralrecon} have proven effective in alleviating the texture-less problem by exploiting various geometric priors learned from a large amount of data. However, these methods struggle to produce high-quality reconstruction of indoor scenes with geometry details. For example, depth-based methods~\cite{teed2018deepv2d,im2019dpsnet} usually estimate depth maps individually, which  causes lack of coherence and scale ambiguities across frames, as well as noisy surface and floating outliers in reconstruction. TSDF-based methods~\cite{murez2020atlas,sun2021neuralrecon} suffer from high memory consumption due to their usage of the explicit 3D volumetric representation; as the memory requirement grows prohibitively large when the resolution is high, they cannot be applied at a level where the fine details can be reconstructed.

Recently, neural scene representations, along with the inverse rendering techniques~\cite{mildenhall2020nerf,wang2021neus,yariv2021volumesdf,yariv2020idr,Oechsle2021unisurf} have shown impressive results on geometry reconstruction, by encoding the volume density, occupancy, or signed distance in a compact and differentiable representation.
However, most neural methods fail to reconstruct indoor scenes with large texture-less regions that do not contain sufficient visual features needed for pixel-level optimization. To address this issue, NerfingMVS~\cite{wei2021nerfingmvs} integrates depth priors to guide the sampling of points in NeRF's framework  \cite{mildenhall2020nerf} to reduce shape-radiance ambiguity. Although it can predict better depth maps than the original NeRF's estimation, the fused geometry of its output depth maps still has limited surface quality.

We present a novel neural surface reconstruction method, called {\em NeuRIS}, that is specialized for indoor scenes. 
Our key idea is to leverage learned normal priors in an adaptive manner to facilitate the learning of the neural surface representation, where the normal priors provide more globally consistent geometric constraints to guide the optimization process.
Specifically, we first estimate the normal maps of the input images using an existing monocular normal estimation network. Then, besides the appearance supervision provided by the input images, the normal priors are used to provide additional constraints to mitigate the geometry ambiguity issue at texture-less regions, which typically consist of smooth or regular shapes.

Note that the normal priors may be inaccurate in the regions with small objects, complex shapes, or thin structures, impeding high-quality reconstruction. 
Hence, for such regions, we propose to use the normal priors in an adaptive manner. To this end, we develop a mechanism to evaluate the faithfulness of the normal priors on-the-fly, based on multi-view photography consistency across the input images. 
For the regions where the multi-view consistency is not satisfied, the normal constraint will be removed and only the appearance information is utilized for optimization.
We observe that the regions where the normal priors are not faithful typically consist of sharp features or irregular shapes with relatively rich visual features in the input images, which are often sufficient for reconstructing high quality surfaces by appearance supervision from the images. 
This adaptive strategy of utilizing the normal priors makes the reconstruction process more robust for general indoor scenes. As a result, NeuRIS achieves high-quality reconstruction of complex indoor scenes with rich geometric details.

To summarize, NeuRIS has the following advantages:
\begin{itemize}
\item We advocate the use of normal priors because they are invariant to translation and scaling, and exhibit better multi-view consistency than the depth prior used in prior methods. The normal priors provide globally consistent geometric constraints across input images, leading to significant improvement of reconstruction quality in texture-less regions of large smooth objects, typically present in indoor scenes. 

\item We apply the normal priors in an adaptive manner, which is achieved by evaluating the faithfulness of the normal priors on-the-fly. This strategy enables complex shapes with geometric details in indoor scenes to be faithfully reconstructed.
\end{itemize}

Extensive validations and comparisons are presented to show that NeuRIS achieves superior results on ScanNet \cite{dai2017scannet} and significantly outperforms the state-of-the-art methods in terms of the reconstruction quality of indoor scenes.

\section{Related works}

\subsection{Indoor scene reconstruction}
\label{subsec:indoor_recons}
Traditional multi-view stereo methods~\cite{schoenberger2016colmap,shen2013openmvs,zheng2014patchmatch} can produce plausible geometry of textured surfaces, but struggle with texture-less regions such as those in indoor scenes. Recently, learning-based MVS methods achieve promising results for tackling texture-less surfaces. Such methods can be divided into two categories: depth based methods~\cite{im2019dpsnet,teed2018deepv2d,long2020cnmnet,long2021multi,long2021B,luo2020cvd} and TSDF (truncated signed distance function) based methods~\cite{murez2020atlas,sun2021neuralrecon}. 
The depth based methods first estimate depth maps of images individually, and then leverage extra filtering and fusion procedures to reconstruct the scene. Such methods often suffer from incompleteness, noisy surfaces and scale ambiguities, due to the inconsistency caused by individual estimation of depth maps. To alleviate these problems, some methods~\cite{murez2020atlas,sun2021neuralrecon}
directly regress input images to TSDF.
Atlas~\cite{murez2020atlas} proposes a volumetric design to regress a 3D global feature volume constructed from a sequence of images to TSDF. Constrained by its global design and computational resources, Atlas can only process a limited number of images, and its reconstruction results lack details.
To reduce the computational burden, unlike Atlas that processes the whole image sequences at once, NeuralRecon~\cite{sun2021neuralrecon} proposes a coarse-to-fine framework, that reconstructs the whole scene by processing local fragments incrementally. However, due to its local estimation design, it is challenging for NeuralRecon to obtain a global reconstruction with fine details.

\subsection{Neural volume rendering and prior guided optimization}

Recently, coordinate-based neural representations, that encode a field by regressing the 3D coordinates to outputting values by Multi-Layer Perceptrons (MLPs), have become a popular way to represent scenes for their compactness and flexibility.
Neural fields have achieved remarkable results on encoding images~\cite{sitzmann2020implicit,chen2021learning,ramasinghe2021beyond}, shapes~\cite{sitzmann2020implicit,park2019deepsdf,gropp2020implicit,atzmon2020sal,mescheder2019occupancy}, and 3D scenes~\cite{mildenhall2020nerf,yariv2020idr,yariv2021volumesdf,Oechsle2021unisurf,wang2021neus,xiangli2021citynerf}. In this paper we mainly focus on neural 3D scene representation and its inverse rendering techniques. Different types of fields are chosen for different goals. Neural Radiance Fields~\cite{mildenhall2020nerf}, which encodes the scene geometry by volume density, is suitable for the tasks of novel view synthesis by volume rendering. However, volume density cannot represent high-fidelity surfaces due to the lack of surface constraints. A better reconstruction of surface geometry can be achieved by using occupancy and signed distances; 
they can be optimized by both surface rendering~\cite{yariv2020idr,niemeyer2020dvr} and volume rendering~\cite{wang2021neus,yariv2021volumesdf,Oechsle2021unisurf} from the supervision of reference images.
In order to further improve the reconstruction accuracy, a concurrent work NeuralWarp~\cite{darmon2021neuralwarp} proposes a warping-based loss term of image patches to improve the reconstruction accuracy. However, these methods perform poorly on indoor scenes because of the lack of textures in indoor scenes. Thus, some methods try to introduce geometric priors to guide the optimization process.

\vspace{1mm}\noindent\textbf{Depth priors.}
Some methods~\cite{wei2021nerfingmvs,liu2020neural} use depth priors to supervise the training process and/or to guide the sampling process of NeRF~\cite{mildenhall2020nerf} for indoor scene rendering to alleviate the shape ambiguity problem. Although they can predict better depth maps than those rendered from NeRF, the inherent problems of depth-based methods described in Sec~\ref{subsec:indoor_recons} still remain, and they cannot produce smooth geometries even after post-processing the data by filtering and fusion. 
Similarly, Roessle \etal~\cite{roessle2021indoorrendering} utilize dense depth priors in NeRF's optimization framework for novel view synthesis with sparse input views. They construct a depth completion network to get dense depth priors from sparse point cloud of SfM. However, the framework is designed for novel view synthesis but not for geometry reconstruction.  Besides, a concurrent work, named MonoSDF \cite{Yu2022MonoSDF}, also integrates learned monocular geometric clues into neural volume rendering framework, where the normal and depth priors are interated into the optimization process to improve the reconstruction quality.

\vspace{1mm}\noindent\textbf{Normal priors.} Surface normal is important for 3D scene understanding \cite{yin2019normalfordepth,zhao2021iterative_normal_depth} and recently single view normal estimation has made great progress with high accuracy~\cite{Do2020tiltedsn,wang2020vplnet,huang2019framenet,Bae2021normaluncertainty}. We observe that the estimated normal priors show high consistency in planar regions and across input views, and also provide obvious clues of underlying geometry, as shown in Fig.~\ref{fig:ab_normal_prior}. 
For the good properties of normal priors, and also in order to avoid the problems brought by depth priors as mentioned above, we choose to integrate normal priors into the volume rendering framework for improving the optimization of the surface representation.

\section{Method}

Given a set of calibrated RGB images $\{\mathcal{I}_k\}$ of an indoor scene, our goal is to accurately reconstruct the scene geometry with fine details. To this end, we adopt a global neural surface representation and optimize it with the supervision of input RGB images. To reconstruct high-fidelity indoor scenes that contain both large texture-less regions and irregular shapes with fine details, we propose an adaptive, prior-guided optimization method. Specifically, we incorporate normal priors learned from a large dataset of indoor scenes into a neural rendering framework for 3D shape reconstruction. Furthermore, noting that normal priors tend to be inaccurate in regions with irregular shapes and thin structures, we propose to use normal priors in an adaptive manner. This is achieved by evaluating the multi-view consistency of the normal priors, so that they are only applied for reconstructing smooth and regular shapes, but not for objects with intricate geometries.

Our pipeline has two phases. In the first phase, we use the normal priors predicted by a monocular method~\cite{Do2020tiltedsn} to provide constraints on the normals rendered with the neural volume rendering framework. Note that the evaluation of the normal prior is not invoked in this phase (Sec~\ref{sec:volume_rendering}). What we obtain in the first phase is a coarse shape with fairly good depth estimations, but lacks local fine details. 
At this stage, large flat shapes are reconstructed in reasonably good quality, thanks to the use of the normal prior, but inaccurate gross shapes are produced for thin structures or small objects with irregular shape features, since the normal priors are not reliable for such areas.
\begin{figure}[t]
  \includegraphics[width=\textwidth]{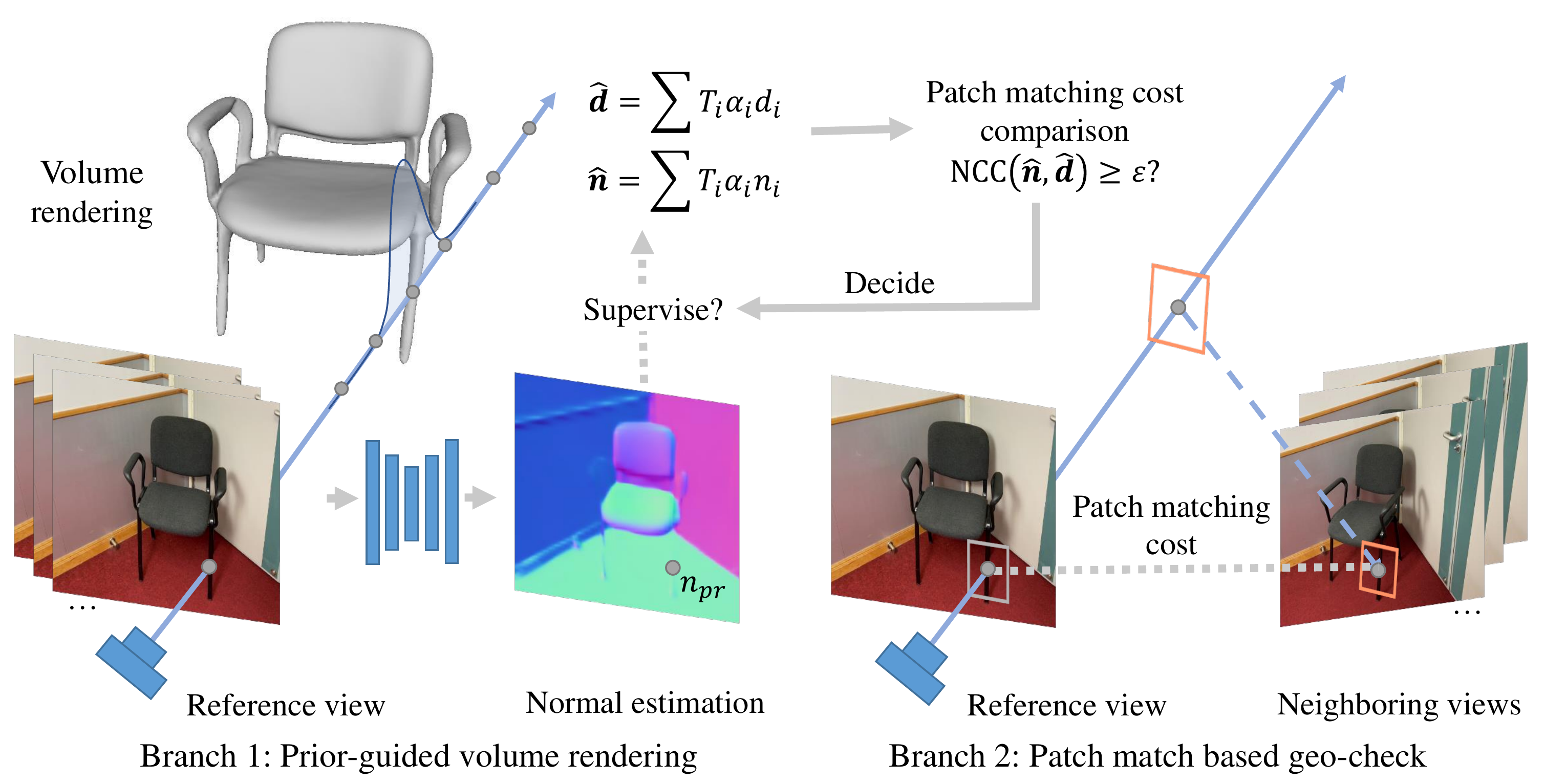}
  \caption{\textbf{Method overview.} Our training process is composed by two phases. In the first phase, we train a coarse model to fit both the multi-view images and the estimated normal maps by volume rendering (Sec~\ref{sec:volume_rendering}), without any filtering strategy. In the second phase, we adaptively impose the supervision from normal priors, where two branches are performed simultaneously: in one branch we conduct a geometric quality evaluation by computing multi-view visual consistency; in the other branch, only those prior normals that pass the geometric check are accepted as proper supervisions to the rendered normals. 
  }
  \label{fig:1_method_overview}
\end{figure}

Hence, in the second phase, we introduce a scheme to evaluate the faithfulness of the normal priors by evaluating the multi-view photometric consistency induced by the currently estimated normals and depths. Only those normal priors whose resulting corresponding geometry passes the photometric-consistency test will be considered reliable and are used for supervision in the following optimization steps. For those unreliable normal priors, we will remove them from the supervision and only rely on the color information 
for the following optimization steps.
This scheme improves the quality of regions where there are sharp geometry features with relatively more visual features (Sec~\ref{sec:patchmatch}). Fig. \ref{fig:1_method_overview} shows an overview of our approach.

\subsection{Prior-guided volume rendering}\label{sec:volume_rendering}

\heading{Scene representation.}
Similar to NeuS~\cite{wang2021neus}, a 3D indoor scene is represented by two Multi-layer Perceptrons (MLPs): geometry network $f_{\theta_g}:\mathbb{R}^3\rightarrow \mathbb{R}$ to encode the signed distance function (SDF), and color network $c_{\theta_c}: \mathbb{R}^3\times\mathbb{R}^3 \rightarrow \mathbb{R}^3$ to encode the colors associated by a spatial position and view direction. The surface $\mathcal{S}$ is then defined as the zero level-set of the SDF, that is,
\begin{equation}
\mathcal{S}=\{\mathbf{x} | f_{\theta_g}(\mathbf{x})=0\}.
\end{equation}

\heading{Volume rendering.} To enable robust supervision using the 2D image observations, we adopt the volume rendering technique, which is proven powerful in NeRF and its variants. Specifically, for each pixel we sample a set of points along the corresponding emitted ray, denoted as $\mathbf{p}_i=\mathbf{o}+d_i\mathbf{v}$, where $\mathbf{p}_i$ are the sampled points, $\mathbf{o}$ is the camera center and $\mathbf{v}$ is the direction of this ray. Then the color is accumulated along the ray through Eq. \ref{eq:volume_rendering}.
\begin{equation}
    \hat{\mathbf{c}} = \sum_{i=1}^{n} T_i \alpha_i c(\mathbf{p}_i, \mathbf{v}),
    \label{eq:volume_rendering}
\end{equation}
where $T_i= \prod_{j=1}^{i-1}(1-\alpha_j)$ denotes the \textit{accumulated transmittance}, $\alpha_i=1-\exp(-\int_{t_i}^{t_{i+1}}\rho(t)dt)$ is the discrete opacity, and the opaque density $\rho(t)$ follows the original definition in NeuS~\cite{wang2021neus}. Since the rendering procedure is fully differentiable, we can learn the weights of $f_{\theta_g}$ and $c_{\theta_c}$ by minimizing the difference between the rendering results and reference images. 

However, as shown in our experiments (Sec~\ref{sec:comparisons}), due to the lack of texture, the pixel-wise colors do not provide sufficient information, thus simply using supervision by the input images would lead to noisy results at texture-less regions.

\vspace{1mm}\heading{Prior-guided optimization.} 
A key observation is that the above volume rendering scheme can generate not only appearance, but also geometric properties such as depth and normal vectors.
That is, we can approximate the surface normal and depth observed from a viewpoint using the volume accumulation along this ray by
\begin{equation}
    \hat{\mathbf{n}} = \sum_{i=1}^{n} T_i \alpha_i \mathbf{n}_i,\ \ 
    \hat{d} = \sum_{i=1}^{n} T_i \alpha_i d_i,
    \label{eq:volume_rendering_normal}
\end{equation}
where $\mathbf{n}_i=\nabla f(\mathbf{p}_i)$ is the spatial gradient of SDF at $\mathbf{p}_i$ and $d_i$ is the corresponding depth.

Given the geometric priors represented by normal maps $\{\mathcal{N}_k\}$ predicted from RGB images $\{\mathcal{I}_k\}$, we supervise the rendered normal $\hat{\mathbf{n}}$ by comparing it with the corresponding estimation from $\mathcal{N}_k$. We use a pre-trained single view normal estimation network~\cite{Do2020tiltedsn} to generate the reference normal maps as supervision. Although the direct use of normal priors without any filtering helps reconstruct complete surfaces, the results still lack fine details. This is because the estimated normal maps are usually over-smoothed, inaccurate even grossly on some delicate structures, such as chair legs, curtains, etc. This motivates us to develop a filtering scheme to use the normal priors in an adaptive manner for reconstructing more accurate surface geometry.

\subsection{Adaptive check of normal priors}\label{sec:patchmatch}
In this section, we introduce a checking method for evaluating the normal quality and adaptively imposing the prior supervision in the optimization process. Our method is developed over a crucial observation: the predicted normal maps are overly smooth in regions where there are sharp geometric details. Moreover, such regions usually have rich visual features, which provide useful clues for validating the accuracy of the normals by evaluating the photometric consistency, i.e., by projecting the reconstructed shape to input images and computing the visual differences across the multi-view images. Based on this observation, we propose a check scheme based on the patch match technique for evaluating the multi-view consistency from rendered depth and normal vectors (Eq. \ref{eq:volume_rendering_normal}). This multi-view consistency evaluation can help NeuRIS identify whether the current geometry is well reconstructed or not. If not, the normal priors would be regarded as unreliable and not used for further refinement of reconstruction.

Specifically, consider evaluating the visual consistency of the surface observed from a pixel $q$ on a reference image $\mathcal{I}_i$, a local 3D plane $\{\mathbf{p} | \mathbf{p}^\intercal\mathbf{n}=d\mathbf{v}^\intercal\mathbf{n}\}$ is defined in the reference camera space associated from $q$, where $\mathbf{v}$ is the view direction, $d$ and $\mathbf{n}$ are the distance and the normal estimation from $q$. We then find a set of neighboring images, and say one of the neighboring images is $\mathcal{I}_j$. The homography transformation from $\mathcal{I}_i$ to $\mathcal{I}_j$ can be computed by Eq. \ref{eq:homography}.

\begin{equation}\label{eq:homography} 
    H_{\mathbf{n},d}=K_j(R_jR_i^{-1}-
    \frac{(\mathbf{t}_i-\mathbf{t}_j)\mathbf{n}^\intercal}
            {d\mathbf{v}^T\mathbf{n}})K_i^{-1}.
\end{equation}
Here $\{K_*, R_*, \mathbf{t}_*\}$ are the camera parameters denoting intrinsic matrices, rotations and translations.

Then for pixel $q$ in $\mathcal{I}_i$, we find a squared patch $P$ centered at it and warp this patch to its neighbor view $\mathcal{I}_j$ with the calculated homography matrix. The visual consistency of $(\mathbf{n}, d)$ is evaluated with the Normalized Cross Correlation (NCC) following the conventional patch-match techniques~\cite{schoenberger2016colmap} by
\begin{equation}\label{eq:ncc}
    {\rm NCC}_j(P,\mathbf{n},d)=\frac{ 
    \sum_{q\in P}\hat{\mathcal{I}}_i(q)\hat{\mathcal{I}}_j(H_{\mathbf{n},d}(q))}
    {
    \sqrt{\sum_{q\in P}\hat{\mathcal{I}}_i(q)^2\sum_{q\in P}\hat{\mathcal{I}}_j(H_{\mathbf{n},d}(q))^2}
    },
\end{equation}
where $\hat{\mathcal{I}}_*(q)=\mathcal{I}_*(q) - \overline{\mathcal{I}}_*(q)$ denotes the result subtracted by the mean value of the local patch.

During the training process, the sampled pixel $q$ along with the patch $P$ in the reference image is associated with the plane by its accumulated depth $\hat{d}$ and normal $\hat{\mathbf{n}}$ in volume rendering. If the reconstructed geometry is not accurate at the sampled pixel, it will fail to satisfy multi-view photometric consistency, which means that its associated normal prior has failed to provide help for the reconstruction process. By comparing the NCC at the sampled patch to a robust threshold $\epsilon$, we can adaptively decide the training weight of normal priors, using the indicator function: 
\begin{equation}
\label{eq:adaptive_weight}
\Omega_q(\hat{\mathbf{n}}, \hat{d}) = \begin{cases}
      1 &  \text{ if } \sum_{j}{\rm NCC}_j(P, \hat{\mathbf{n}}, \hat{d}) \geq \epsilon\\
      0 & \text{ if } \sum_{j}{\rm NCC}_j(P, \hat{\mathbf{n}}, \hat{d}) < \epsilon. 
      \end{cases}   
\end{equation}
Only when $\Omega_q (\hat{\mathbf{n}}, \hat{d})$ equals $1$ the normal prior will be used for supervision. And once the normal priors are judged as unfaithful, they will not be used in the subsequent optimization process.

\subsection{Training}
In the training stage, we sample a batch of pixels and adaptively minimize the difference of the color and normal estimations and the corresponding references. Specifically, during training, in each iteration we sample $m$ pixels $\{q_k\}$ and their corresponding reference colors $\{\mathcal{I}(q_k)\}$ and normals $\{\mathcal{N}(q_k)\}$. For each pixel we sample $n$ points along its corresponding ray in the world space. The overall loss is defined as 

\begin{equation}
    \mathcal{L} = \lambda_c\mathcal{L}_c + 
    \lambda_{p} \mathcal{L}_p + 
    \lambda_{eik} \mathcal{L}_{eik}.
\end{equation}
Here the color loss $\mathcal{L}_{c}$ is defined as
\begin{equation}
    \mathcal{L}_{c} = \frac{1}{m}\sum_{k}\|\mathcal{I}(q_k)-\hat{\mathbf{c}}(q_k)\|_1,
\end{equation}
where $\hat{\mathbf{c}}(q_k)$ is the predicted pixel colors by volume rendering. The normal prior loss is denoted by
\begin{equation}
    \mathcal{L}_{p} = \frac{1}{m}\sum_{k}\|\mathcal{N}(q_k)-\hat{\mathbf{n}}(q_k)\|_1\cdot \Omega_{q_k}(\hat{\mathbf{n}}(q_k),\hat{d}(q_k)).
\end{equation}
Note that there are two phases in the whole training process and at the first phase, there is no geometric check. Thus, the indicator $\Omega_{q_k}(\hat{\mathbf{n}}(q_k),\hat{d}(q_k))$ is always equal to 1 at the first phase while it follows Eq.~\ref{eq:adaptive_weight} at the second phase.

The Eikonal loss~\cite{icml2020igr} to regularize the SDF is defined as
\begin{equation}
    \mathcal{L}_{eik} = \frac{1}{nm}\sum_{k,i}(\|\nabla f(\mathbf{p}_{k,i})\|_2-1)^2.
\end{equation}
$\lambda_c, \lambda_p, \lambda_{eik}$ are hyperparameters for weighting color loss, prior loss and Eikonal loss, respectively.

\section{Experiments}
\subsection{Implementation details}
\textbf{Architecture.} We adopt the same network architecture of NeuS~\cite{wang2021neus}, where the signed distance function and color function are modeled by an MLP with 8 and 6 hidden layers respectively. Positional encoding~\cite{mildenhall2020nerf} and sphere initialization~\cite{atzmon2020sal} are applied to the network. For the normal priors, we adopt the recent method~\cite{Do2020tiltedsn} and re-train its network with our training/test splits to predict the normals of input images  instead of using its officially pretrained model. We sample 512 rays for each batch to train the model. And we first train the model for 60k iterations with normal priors and then continue to train the full model for another 100k iterations, which takes about 10 hours in total on a single NVIDIA RTX2080Ti GPU. More details can be found in the supplementary.

\vspace{1mm}\noindent\textbf{Dataset.} We test the performance of our algorithm on ScanNet~\cite{dai2017scannet}. ScanNet is a large-scale dataset consisting of 1613 indoor scenes with ground truth camera intrinsics, camera poses and surface reconstructions. Following NerfingMVS~\cite{wei2021nerfingmvs}, we randomly select 8 scenes and all images are resized in $640\times 480$ resolution. Different from NerfingMVS~\cite{wei2021nerfingmvs} using images covering a local region in a room, we aim to perform room-scale reconstruction. For each scene, a set of equally-spaced images (about 150$\sim $600 images) is sampled from the corresponding video, thus the number of sampled images is proportional to the video length.

\vspace{1mm}\noindent\textbf{Baselines.} 
We compare our method with the following methods: (1) Depth-based method: DeepV2D~\cite{teed2018deepv2d}; (2) TSDF based methods: NeuralRecon~\cite{sun2021neuralrecon} and Atlas~\cite{murez2020atlas}; (3) Neural volume rendering methods: NeRF\cite{mildenhall2020nerf}, NerfingMVS ~\cite{wei2021nerfingmvs}, NeuS~\cite{wang2021neus} and VolSDF~\cite{yariv2021volumesdf}; and (4) Traditional MVS reconstruction method COLMAP~\cite{schoenberger2016colmap}. For the depth based method DeepV2D, to address the scale ambiguity issue of it,
we re-scale every predicted depth map according to the ground truth depth map using the median scale strategy~\cite{zhou2017scalemedian} and then fuse its predicted depth maps following NeuralRecon~\cite{sun2021neuralrecon} to construct global surface geometry.
For COLMAP, we use ground truth poses to reconstruct point cloud and then use Screened Poisson Surface Reconstruction~\cite{kazhdan2013poissonrecon} to get a mesh. 

\vspace{1mm}\noindent\textbf{Evaluation metrics.}
For complete quantitative comparisons, we evaluate the 3D surface geometry results, following the metrics defined in NeuralRecon~\cite{sun2021neuralrecon}. Among those metrics, F-score is usually considered as the most suitable metric to evaluate geometry quality ~\cite{sun2021neuralrecon}. Refer to the supplementary for more details.

\subsection{Comparisons}
\label{sec:comparisons}
\textbf{3D reconstruction.} Table~\ref{tab:evaluation_mesh} shows the quantitative results compared with the state-of-the-art methods.  Note that for the data-driven methods~\cite{sun2021neuralrecon,murez2020atlas,teed2018deepv2d}, we use the official pre-trained models. As shown in Table~\ref{tab:evaluation_mesh}, our method can significantly surpass existing methods, especially when compared with neural volume rendering methods.
For the metric Comp. (Completeness), NeuRIS is slightly worse than DeepV2D. This is because we scale each depth map of DeepV2D to the GT depth map, and thus after the fusion of the depth maps, there are sufficient points near the GT mesh (i.e., a low Comp. error), but also many points are still far away from the GT mesh (i.e., a high Accuracy error). For NerfingMVS~\cite{wei2021nerfingmvs}, it is designed for refining depth maps, thus we fuse its output depth maps through TSDF fusion~\cite{newcombe2011kinectfusion}. Note that it failed on most (5/8) room-scale scenes in our pre-experiments and it only showed the results of reconstructing local room regions in their original paper. Thus, we only report the averaged scores on succeeded scenes here. For NeRF, we use the level set 20 to extract surfaces, where the level set is carefully selected (See the supplementary). 

Fig.~\ref{fig:comparison} shows the qualitative comparisons. Our method is visually much better than other methods with fine details. We remark that our method can produce much complete and smooth results and fill the holes that exist in the ground truth surface, which is mainly caused by occlusions and incomplete scans~\cite{murez2020atlas}. Refer to the supplementary for more qualitative results on Scannet and other indoor datasets.

\begin{table}
    \centering
    \caption{{\bf Quantitative comparisons of room-scale reconstruction results} over 8 scenes using 3D geometry metrics. For VolSDF and NerfingMVS, the scores are averaged on 5 and 3 scenes respectively because they failed on other scenes.}
    \begin{center}
    \begin{tabular}{>{\centering\arraybackslash}p{3cm} >{\centering\arraybackslash}p{1.5cm} >{\centering\arraybackslash}p{1.5cm} >{\centering\arraybackslash}p{1.5cm} >{\centering\arraybackslash}p{1.5cm} >{\centering\arraybackslash}p{1.5cm} >{\centering\arraybackslash}p{1.5cm}}
    \hline\noalign{\smallskip}
     Method & Accu. $\downarrow $ & Comp.$\downarrow $  & Prec.$ \uparrow $ &   Recall$\uparrow $ & F-score$\uparrow $  \\
    \hline\noalign{\smallskip}
    COLMAP\cite{schoenberger2016colmap} & 0.076  &   0.096  &   0.559  &   0.545  &   0.548\\
     NeuralRecon\cite{sun2021neuralrecon} &  0.046  &   0.081  &   0.720  &   0.577  &   0.640\\
    Atlas\cite{murez2020atlas}  & 0.211  &   0.070  &   0.500  &   0.659  &   0.564 \\
    DeepV2D\cite{teed2018deepv2d} &  0.174  & \textbf{0.049}  &   0.528  &   0.682  &   0.593 \\
    NeRF\cite{mildenhall2020nerf} &   0.127  &   0.080  &   0.404  &   0.512  &   0.436 \\
    NerfingMVS\cite{wei2021nerfingmvs}    &   0.155  &   0.087  &   0.410  &   0.471  &   0.438 \\
    NeuS\cite{wang2021neus} & 0.183  &   0.152  &   0.286  &   0.290  &   0.284 \\
    VolSDF\cite{yariv2021volumesdf} & 0.237  &   0.171  &   0.331  &   0.280  &   0.301 \\
    Ours &  \textbf{0.046}    &   0.053  &   \textbf{0.770}  &   \textbf{ 0.707}  &   \textbf{0.736} \\
    \hline
    \end{tabular}
    \end{center}
    \label{tab:evaluation_mesh}
\end{table}
\begin{figure}
    \centering
    \includegraphics[width=\textwidth]{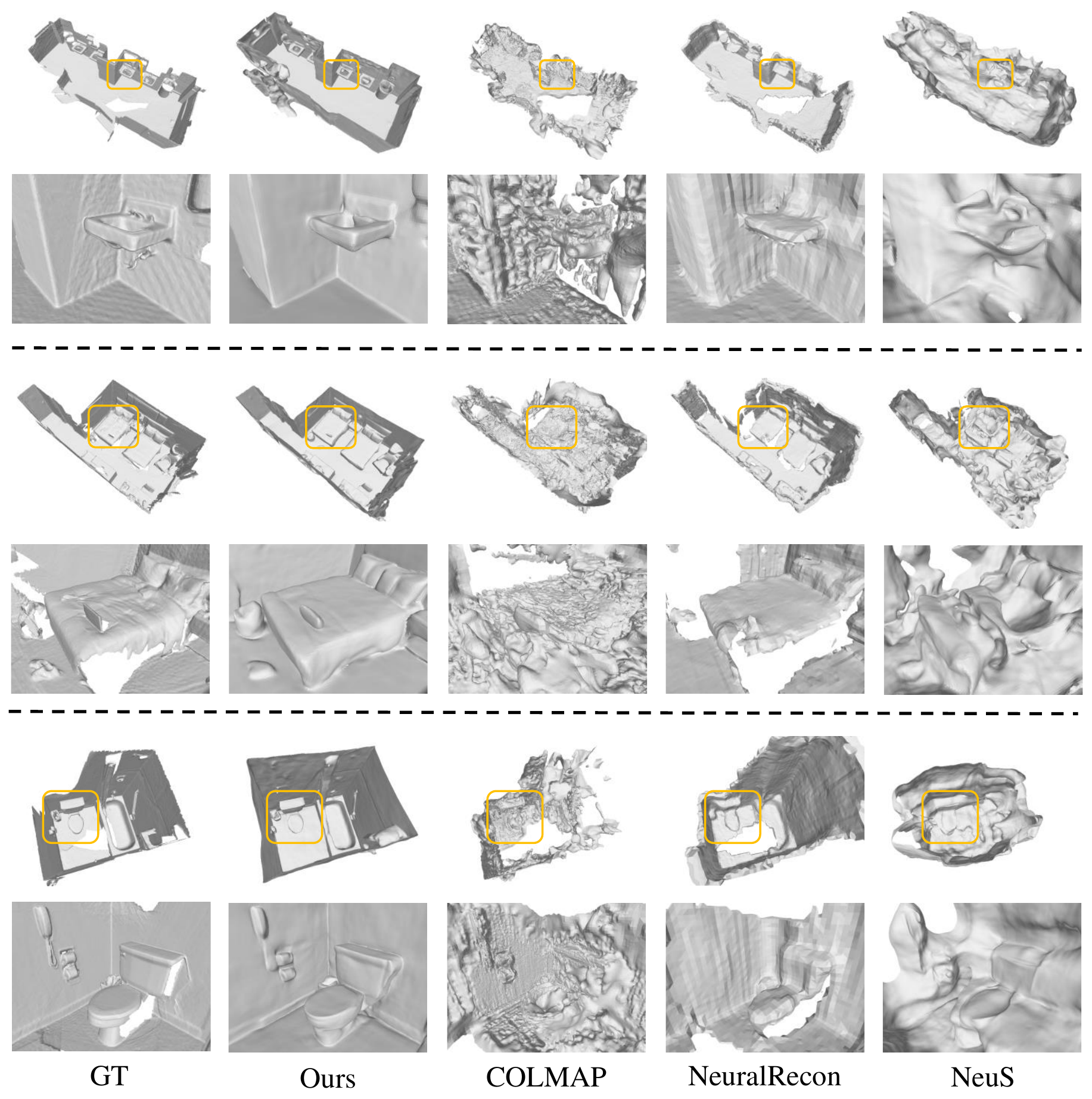}
    \caption{\textbf{Qualitative geometry comparisons.} For each block, the first row is the top view of the whole room and the second row is the zoom-in view of the marked area. Our method can produce more accurate and complete reconstruction results and preserve fine details of the scenes.}
    \vspace{-8pt}
    \label{fig:comparison}
\end{figure}
\begin{figure}
  \centering
  \includegraphics[width=0.9\linewidth]{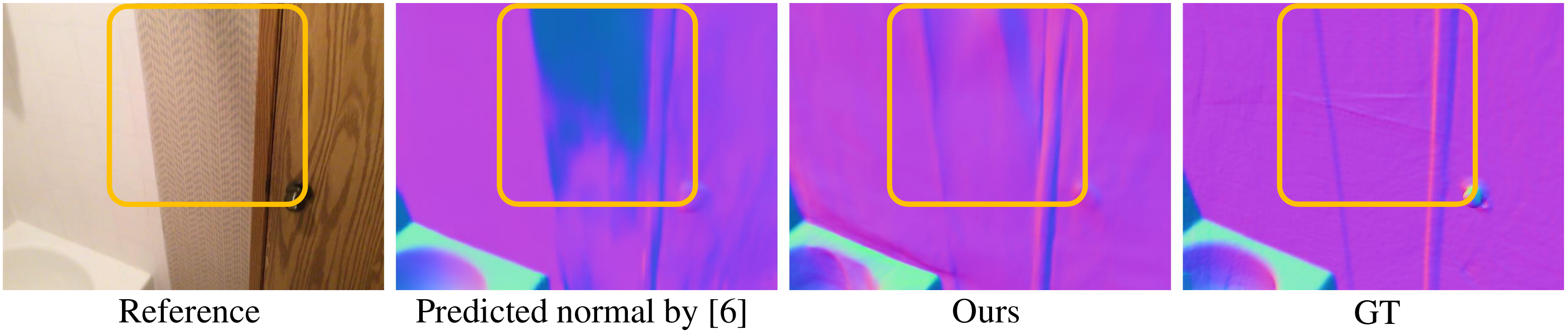}
   \caption{\textbf{Qualitative normal comparisons.} When the wall is observed too locally, the normal estimation of [6] in the marked area is not accurate while it can be improved by leveraging multi-view information with better observations in our method. 
   }
   \label{fig:evaluation_normal}
\end{figure}

\vspace{1mm}\noindent\textbf{Normal predictions.} Except for accurate geometry reconstruction, our method can also achieve more accurate normal predictions than \cite{Do2020tiltedsn}. For the monocular normal estimation method \cite{Do2020tiltedsn}, the estimated normal maps may contain wrong predictions when there are severe occlusions or ambiguities. For example, when a wall is observed too locally, 
it's hard to utilize the global information for precise normal estimations. Thanks to the proposed adaptive normal guided optimization, our method can improve the global consistency of normal maps across views and correct the wrong estimations from \cite{Do2020tiltedsn}. Fig. \ref{fig:evaluation_normal} clearly shows one example that our method can improve the quality of normal estimations. 
Quantitative comparisons summarized in Table \ref{tab:evaluation_normal} also validates NeuRIS's normals are better than those from \cite{Do2020tiltedsn}, using the metrics defined in \cite{Do2020tiltedsn}. Here, we compare the cosine similarity of our rendered normal (Eq. \ref{eq:volume_rendering_normal}) and the predicted normal of \cite{Do2020tiltedsn} with the GT normal over 8 scenes (i.e., 493 images), respectively.

\begin{table}[h]
    \centering
    \caption{\textbf{Quantitative normal evaluation} over 8 scenes.}
        \begin{tabular}{>{\centering\arraybackslash}p{2cm} >{\centering\arraybackslash}p{1.2cm} >{\centering\arraybackslash}p{1.3cm} >{\centering\arraybackslash}p{1.3cm} >{\centering\arraybackslash}p{1.3cm} >{\centering\arraybackslash}p{1.2cm} >{\centering\arraybackslash}p{1.2cm}}
        \hline\noalign{\smallskip}
         Method & Mean$\downarrow $ & Median$\downarrow $  &RMSE$\downarrow$ &  11.25$^{\circ}\uparrow$ & 22.5$^{\circ}\uparrow $ & 30$^{\circ}\uparrow $  \\
        \hline\noalign{\smallskip}
        TiltedSN [6] & 15.4 & 7.3 & 24.8 & 63.2 & 79.3 & 84.5 \\
        Ours  & \textbf{14.7} & \textbf{6.9} & \textbf{24.3} & \textbf{65.4} & \textbf{81.1} & \textbf{86.0}\\
        \hline
        \end{tabular}
    \label{tab:evaluation_normal}
\end{table}

\begin{figure}
  \centering
  \includegraphics[width=\linewidth]{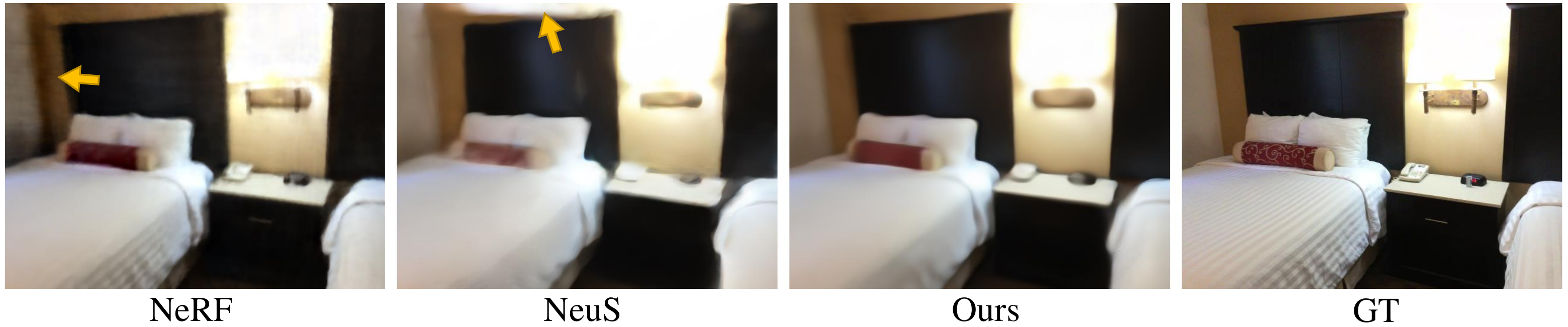}
   \caption{\textbf{Novel view synthesis results.} Our method can produce much better rendering results than the baseline methods NeRF and NeuS.}
   \label{fig:evaluation_nvs}
\end{figure}

\vspace{1mm}\noindent\textbf{Novel view synthesis.} To evaluate the quality of novel view synthesis, we uniformly sample 500 novel views over 8 scenes, which are different from training images. Our rendering quality is better than those of NeRF and NeuS, benefitting from our high-quality geometry. The average PSNR of NeRF, NeuS and ours are 23.3, 22.7 and 24.4, respectively. Fig. \ref{fig:evaluation_nvs} shows one sample of qualitative comparisons and refer to the supplementary for more results.

\subsection{Ablation Studies}
\begin{figure}[h]
  \vspace{-16pt}
  \includegraphics[width=\textwidth]{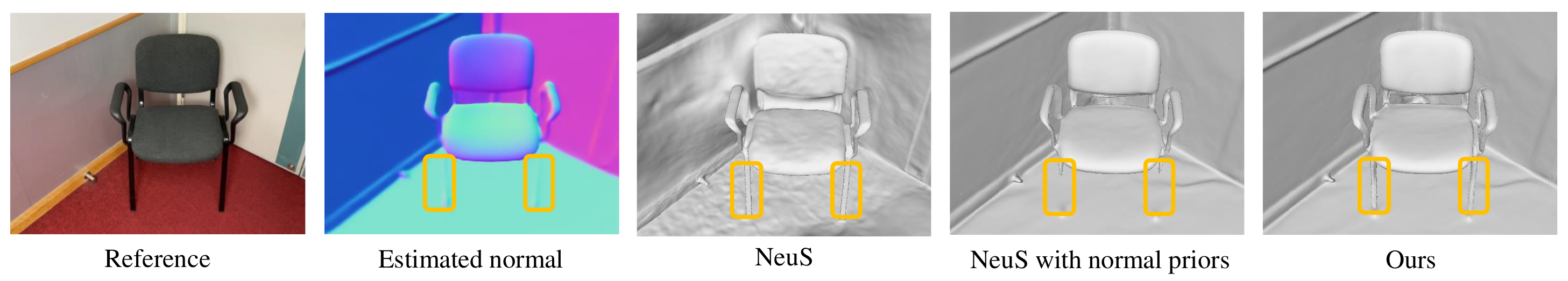}
  \caption{{\bf Ablation study.} The estimated normal priors of reference images show high fidelity at planar regions but they are not correct at chair legs. 
  Naively using all normal priors as supervision can help reconstruct the planar regions but fail to reconstruct the chair legs, while our method can reconstruct them all in high quality.}
  \label{fig:ab_normal_prior}
  \vspace{-8pt}
\end{figure}

In order to evaluate the effectiveness of our proposed components, we conduct experiments in three different settings: (1) NeuS with the default setting; (2) NeuS with normal priors; (3) Ours: NeuS with normal priors and geometric check. Table~\ref{tab:ablation_study} shows that integrating normal priors significantly improves the reconstruction quality because it reduces ambiguities caused by lack of texture. With geometry-check, we can remove the wrongly estimated normals and further improve geometry quality. Moreover, as shown in Figure \ref{fig:ab_normal_prior}, adopting all normal priors naively in NeuS can reconstruct the wall and floor but fail to reconstruct the chair leg. For the pixels corresponding to the chair leg, their multi-view consistency constraints are not satisfied and the normal priors at this area should be removed during the training process. Finally with our geometry check the chair leg can be successfully reconstructed. This demonstrates that our geometry-check can remove wrongly estimated normals.

\begin{table}
    \vspace{-8pt}
    \centering
    \caption{Ablation studies of each component of our method over 8 scenes.}
    \begin{tabular}{>{\centering\arraybackslash}p{0.9cm} >{\centering\arraybackslash}p{0.9cm} >{\centering\arraybackslash}p{0.9cm} >{\centering\arraybackslash}p{1.4cm} >{\centering\arraybackslash}p{1.4cm} >{\centering\arraybackslash}p{1.4cm} >{\centering\arraybackslash}p{1.2cm}
    >{\centering\arraybackslash}p{1.4cm}}
    \hline\noalign{\smallskip}
    NeuS & Prior & Geo  & Accu.$\downarrow $ & Comp.$ \downarrow $  & Prec.$ \uparrow $ &   Recall$\uparrow $ & F-score$\uparrow $ \\
    \noalign{\smallskip}
    \hline\noalign{\smallskip}
    \checkmark &              &              &                  0.183  &   0.152  &   0.286  &   0.290  &   0.284  \\
    \checkmark &  \checkmark  &              &                  0.050  &   0.053  &   0.749  &   0.701  &   0.724  \\
    \checkmark &  \checkmark  &   \checkmark &                  \textbf{0.046}  &   \textbf{0.053}  &   \textbf{0.770}  & \textbf{0.707}  & \textbf{ 0.736} \\
    \hline
    \end{tabular}
    \label{tab:ablation_study}
    \vspace{-12pt}
\end{table}

\begin{figure}
  \includegraphics[width=\textwidth]{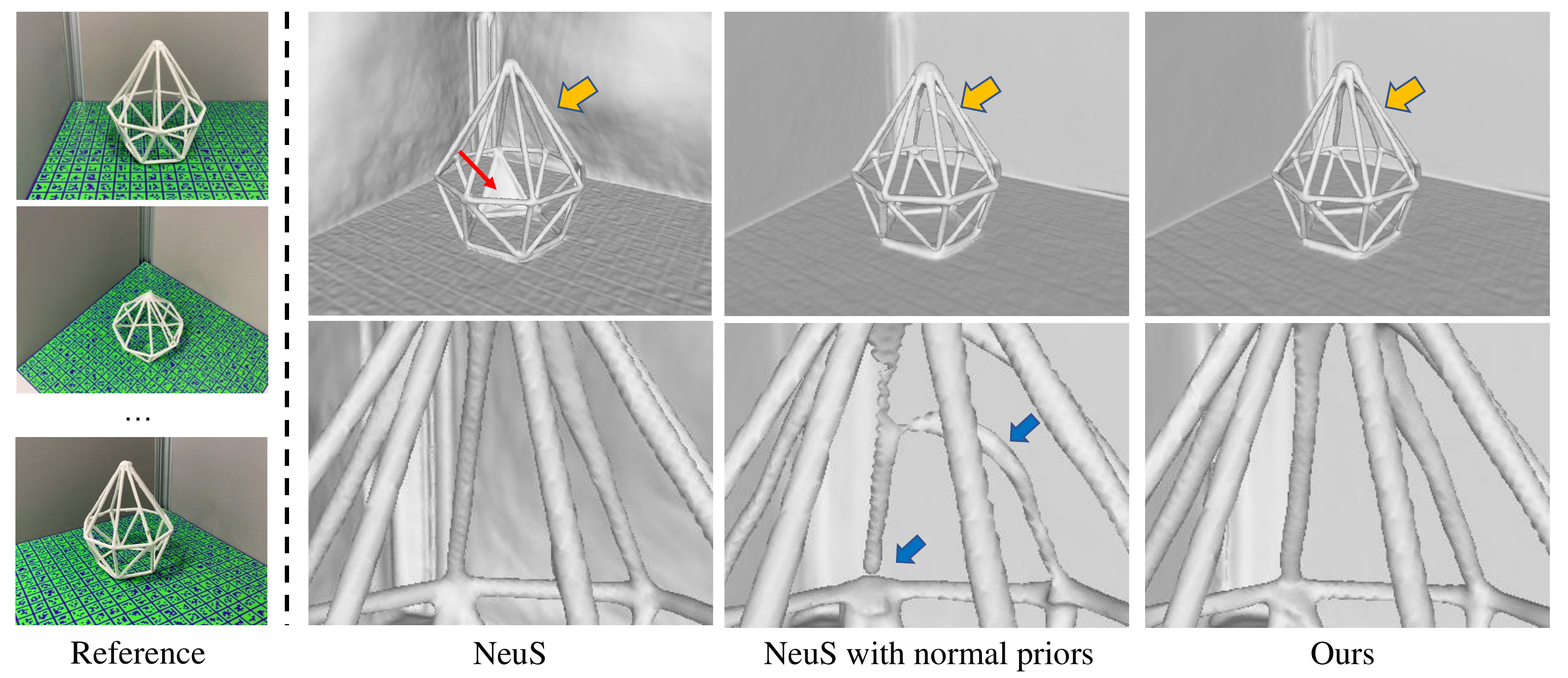}
  \caption{{\bf Reconstruction of a thin structure and desk corner.} The left part shows the reference images. The right part shows the reconstruction results under 3 different settings, where the first row shows the overview of reconstruction results and the second row shows the zoom-in view of the area indicated by yellow arrows.}
  \label{fig:ab_thin_structure}
  \vspace{-4pt}
\end{figure}

We also show the reconstruction results of a challenging thin structure put at a desk corner, given a set of images sampled from a video sequence. 
Fig.~\ref{fig:ab_thin_structure} shows that NeuS can reconstruct the thin structure but there are artifacts, including wrongly reconstructed desk surfaces and redundant surfaces indicated by the red arrow. NeuS with normal priors can reconstruct the background desk surface well but still fail to reconstruct some parts of the thin structure well (blue arrows). As for our method, both the background desk surfaces and foreground thin structures can be well reconstructed. Moreover, different from the methods~\cite{Tabb:cvpr:2013,Liu:sigg:2017,vid2curve,Liu:sigga:2018} which only focus on thin structure reconstruction, our method does not need foreground extraction for each input image as preprocessing and can handle hybrid scenes which contain both thin structures and general objects. 

In summary, for indoor scenes, the predominant classes of shapes are planar or regular surfaces, such as walls, floors, and furniture, which normally possess weak or no visual texture but occupy most areas of the space. With the help of normal priors, these shapes can be well reconstructed. Thus, we can see that the reconstruction quality can be greatly improved. However, normal priors may be not accurate at relatively small objects or thin structures, such as object edges and chair legs. With our geo-check mechanism, these wrongly estimated normals can be removed, therefore the reconstruction quality can be further improved. Although the areas occupied by edges or small objects may not be large in a room, the accurate reconstruction of them is critical to the overall success of reconstruction in terms of perceptual quality.

\section{Conclusion and future work}
In this work, we propose a novel prior-guided optimization framework of neural volume rendering with geometric constraints, which can adaptively integrate normal priors into neural volume rendering efficiently and accurately. This way enables the network to utilize prior knowledge at texture-less areas and maintain the capacity to reconstruct fine details of small objects with relatively more texture. This method has practical uses in VR/AR or other applications that require precise indoor geometry. 
Currently, our method requires per-scene optimization for several hours, which hinders our method from reconstructing scenes at a very large scale. In the future, we will try to integrate some hybrid neural representations such as the multi-resolution hash encoding~\cite{mueller2022instant-ngp} into our model to speed up the training process and also try to adaptively integrate other kinds of priors into our framework, such as plane priors and depth priors, to get better reconstruction quality.

\section*{Acknowlegements}
We thank Yuan Liu and Nenglun Chen for the help with experiments. Christian Theobalt was supported by ERC Consolidator Grant 770784. Lingjie Liu was supported by Lise Meitner Postdoctoral Fellowship. Computational resources are mainly provided by HKU GPU Farm.



%
%
\bibliographystyle{splncs04}

\begin{thebibliography}{10}
\providecommand{\url}[1]{\texttt{#1}}
\providecommand{\urlprefix}{URL }
\providecommand{\doi}[1]{https://doi.org/#1}

\bibitem{atzmon2020sal}
Atzmon, M., Lipman, Y.: Sal: Sign agnostic learning of shapes from raw data.
  In: Proceedings of the IEEE/CVF Conference on Computer Vision and Pattern
  Recognition. pp. 2565--2574 (2020)

\bibitem{Bae2021normaluncertainty}
Bae, G., Budvytis, I., Cipolla, R.: Estimating and exploiting the aleatoric
  uncertainty in surface normal estimation. In: International Conference on
  Computer Vision (ICCV) (2021)

\bibitem{chen2021learning}
Chen, Y., Liu, S., Wang, X.: Learning continuous image representation with
  local implicit image function. In: Proceedings of the IEEE/CVF Conference on
  Computer Vision and Pattern Recognition. pp. 8628--8638 (2021)

\bibitem{dai2017scannet}
Dai, A., Chang, A.X., Savva, M., Halber, M., Funkhouser, T., Nie{\ss}ner, M.:
  Scannet: Richly-annotated 3d reconstructions of indoor scenes. In: Proc.
  Computer Vision and Pattern Recognition (CVPR), IEEE (2017)

\bibitem{darmon2021neuralwarp}
Darmon, F., Bascle, B., Devaux, J.C., Monasse, P., Aubry, M.: Improving neural
  implicit surfaces geometry with patch warping. arXiv preprint
  arXiv:2112.09648  (2021)

\bibitem{Do2020tiltedsn}
Do, T., Vuong, K., Roumeliotis, S.I., Park, H.S.: Surface normal estimation of
  tilted images via spatial rectifier. In: Proc. of the European Conference on
  Computer Vision. Virtual Conference (August 23--28 2020)

\bibitem{gropp2020implicit}
Gropp, A., Yariv, L., Haim, N., Atzmon, M., Lipman, Y.: Implicit geometric
  regularization for learning shapes. arXiv preprint arXiv:2002.10099  (2020)

\bibitem{icml2020igr}
Gropp, A., Yariv, L., Haim, N., Atzmon, M., Lipman, Y.: Implicit geometric
  regularization for learning shapes. In: Proceedings of Machine Learning and
  Systems 2020, pp. 3569--3579 (2020)

\bibitem{huang2019framenet}
Huang, J., Zhou, Y., Funkhouser, T., Guibas, L.J.: Framenet: Learning local
  canonical frames of 3d surfaces from a single rgb image. In: Proceedings of
  the IEEE/CVF International Conference on Computer Vision. pp. 8638--8647
  (2019)

\bibitem{im2019dpsnet}
Im, S., Jeon, H.G., Lin, S., Kweon, I.S.: Dpsnet: End-to-end deep plane sweep
  stereo. arXiv preprint arXiv:1905.00538  (2019)

\bibitem{kazhdan2013poissonrecon}
Kazhdan, M., Hoppe, H.: Screened poisson surface reconstruction. ACM
  Transactions on Graphics (ToG)  \textbf{32}(3),  1--13 (2013)

\bibitem{Liu:sigg:2017}
Liu, L., Ceylan, D., Lin, C., Wang, W., Mitra, N.J.: Image-based reconstruction
  of wire art  \textbf{36}(4),  63:1--63:11 (2017)

\bibitem{Liu:sigga:2018}
Liu, L., Chen, N., Ceylan, D., Theobalt, C., Wang, W., Mitra, N.J.:
  Curvefusion: Reconstructing thin structures from rgbd sequences
  \textbf{37}(6) (2018)

\bibitem{liu2020neural}
Liu, L., Gu, J., Lin, K.Z., Chua, T.S., Theobalt, C.: Neural sparse voxel
  fields. NeurIPS  (2020)

\bibitem{long2021B}
Long, X., Lin, C., Liu, L., Li, W., Theobalt, C., Yang, R., Wang, W.: Adaptive
  surface normal constraint for depth estimation. ICCV  (2021)

\bibitem{long2021multi}
Long, X., Liu, L., Li, W., Theobalt, C., Wang, W.: Multi-view depth estimation
  using epipolar spatio-temporal network. CVPR  (2021)

\bibitem{long2020cnmnet}
Long, X., Liu, L., Theobalt, C., Wang, W.: Occlusion-aware depth estimation
  with adaptive normal constraints. In: European Conference on Computer Vision.
  pp. 640--657. Springer (2020)

\bibitem{luo2020cvd}
Luo, X., Huang, J., Szeliski, R., Matzen, K., Kopf, J.: Consistent video depth
  estimation  \textbf{39}(4) (2020)

\bibitem{mescheder2019occupancy}
Mescheder, L., Oechsle, M., Niemeyer, M., Nowozin, S., Geiger, A.: Occupancy
  networks: Learning 3d reconstruction in function space. In: Proceedings of
  the IEEE/CVF Conference on Computer Vision and Pattern Recognition. pp.
  4460--4470 (2019)

\bibitem{mildenhall2020nerf}
Mildenhall, B., Srinivasan, P.P., Tancik, M., Barron, J.T., Ramamoorthi, R.,
  Ng, R.: Nerf: Representing scenes as neural radiance fields for view
  synthesis. In: European conference on computer vision. pp. 405--421. Springer
  (2020)

\bibitem{mueller2022instant-ngp}
M\"uller, T., Evans, A., Schied, C., Keller, A.: Instant neural graphics
  primitives with a multiresolution hash encoding. arXiv:2201.05989  (Jan 2022)

\bibitem{murez2020atlas}
Murez, Z., van As, T., Bartolozzi, J., Sinha, A., Badrinarayanan, V.,
  Rabinovich, A.: Atlas: End-to-end 3d scene reconstruction from posed images.
  In: ECCV (2020), \url{https://arxiv.org/abs/2003.10432}

\bibitem{niemeyer2020dvr}
Niemeyer, M., Mescheder, L., Oechsle, M., Geiger, A.: Differentiable volumetric
  rendering: Learning implicit 3d representations without 3d supervision. In:
  Proceedings of the IEEE/CVF Conference on Computer Vision and Pattern
  Recognition. pp. 3504--3515 (2020)

\bibitem{Oechsle2021unisurf}
Oechsle, M., Peng, S., Geiger, A.: Unisurf: Unifying neural implicit surfaces
  and radiance fields for multi-view reconstruction. In: International
  Conference on Computer Vision (ICCV) (2021)

\bibitem{park2019deepsdf}
Park, J.J., Florence, P., Straub, J., Newcombe, R., Lovegrove, S.: Deepsdf:
  Learning continuous signed distance functions for shape representation. In:
  Proceedings of the IEEE/CVF Conference on Computer Vision and Pattern
  Recognition. pp. 165--174 (2019)

\bibitem{ramasinghe2021beyond}
Ramasinghe, S., Lucey, S.: Beyond periodicity: Towards a unifying framework for
  activations in coordinate-mlps. arXiv preprint arXiv:2111.15135  (2021)

\bibitem{roessle2021indoorrendering}
Roessle, B., Barron, J.T., Mildenhall, B., Srinivasan, P.P., Nie{\ss}ner, M.:
  Dense depth priors for neural radiance fields from sparse input views. arXiv
  preprint arXiv:2112.03288  (2021)

\bibitem{schoenberger2016colmap}
Sch\"{o}nberger, J.L., Zheng, E., Pollefeys, M., Frahm, J.M.: Pixelwise view
  selection for unstructured multi-view stereo. In: European Conference on
  Computer Vision (ECCV) (2016)

\bibitem{shen2013openmvs}
Shen, S.: Accurate multiple view 3d reconstruction using patch-based stereo for
  large-scale scenes. IEEE transactions on image processing  \textbf{22}(5),
  1901--1914 (2013)

\bibitem{sitzmann2020implicit}
Sitzmann, V., Martel, J.N., Bergman, A.W., Lindell, D.B., Wetzstein, G.:
  Implicit neural representations with periodic activation functions. arXiv
  preprint arXiv:2006.09661  (2020)

\bibitem{sun2021neuralrecon}
Sun, J., Xie, Y., Chen, L., Zhou, X., Bao, H.: {NeuralRecon}: Real-time
  coherent {3D} reconstruction from monocular video. CVPR  (2021)

\bibitem{Tabb:cvpr:2013}
Tabb, A.: Shape from silhouette probability maps: Reconstruction of thin
  objects in the presence of silhouette extraction and calibration error. pp.
  161--168 (June 2013). \doi{10.1109/CVPR.2013.28}

\bibitem{teed2018deepv2d}
Teed, Z., Deng, J.: Deepv2d: Video to depth with differentiable structure from
  motion. arXiv preprint arXiv:1812.04605  (2018)

\bibitem{shen2018mvdepthnet}
Wang, K., Shen, S.: Mvdepthnet: real-time multiview depth estimation neural
  network. In: International Conference on 3D Vision (3DV) (Sep 2018)

\bibitem{vid2curve}
Wang, P., Liu, L., Chen, N., Chu, H.K., Theobalt, C., Wang, W.: Vid2curve:
  Simultaneous camera motion estimation and thin structure reconstruction from
  an rgb video. ACM Trans. Graph.  \textbf{39}(4) (Jul 2020)

\bibitem{wang2021neus}
Wang, P., Liu, L., Liu, Y., Theobalt, C., Komura, T., Wang, W.: Neus: Learning
  neural implicit surfaces by volume rendering for multi-view reconstruction.
  arXiv preprint arXiv:2106.10689  (2021)

\bibitem{wang2020vplnet}
Wang, R., Geraghty, D., Matzen, K., Szeliski, R., Frahm, J.M.: Vplnet: Deep
  single view normal estimation with vanishing points and lines. In:
  Proceedings of the IEEE/CVF Conference on Computer Vision and Pattern
  Recognition. pp. 689--698 (2020)

\bibitem{wei2021nerfingmvs}
Wei, Y., Liu, S., Rao, Y., Zhao, W., Lu, J., Zhou, J.: Nerfingmvs: Guided
  optimization of neural radiance fields for indoor multi-view stereo. In: ICCV
  (2021)

\bibitem{xiangli2021citynerf}
Xiangli, Y., Xu, L., Pan, X., Zhao, N., Rao, A., Theobalt, C., Dai, B., Lin,
  D.: Citynerf: Building nerf at city scale. arXiv preprint arXiv:2112.05504
  (2021)

\bibitem{yariv2021volumesdf}
Yariv, L., Gu, J., Kasten, Y., Lipman, Y.: Volume rendering of neural implicit
  surfaces. Advances in Neural Information Processing Systems  \textbf{34}
  (2021)

\bibitem{yariv2020idr}
Yariv, L., Kasten, Y., Moran, D., Galun, M., Atzmon, M., Ronen, B., Lipman, Y.:
  Multiview neural surface reconstruction by disentangling geometry and
  appearance. Advances in Neural Information Processing Systems  \textbf{33}
  (2020)

\bibitem{yin2019normalfordepth}
Yin, W., Liu, Y., Shen, C., Yan, Y.: Enforcing geometric constraints of virtual
  normal for depth prediction. In: Proceedings of the IEEE/CVF International
  Conference on Computer Vision. pp. 5684--5693 (2019)

\bibitem{Yu2022MonoSDF}
Yu, Z., Peng, S., Niemeyer, M., Sattler, T., Geiger, A.: Monosdf: Exploring
  monocular geometric cues for neural implicit surface reconstruction.
  arXiv:2022.00665  (2022)

\bibitem{zhao2021iterative_normal_depth}
Zhao, W., Liu, S., Wei, Y., Guo, H., Liu, Y.J.: A confidence-based iterative
  solver of depths and surface normals for deep multi-view stereo. In:
  Proceedings of the IEEE/CVF International Conference on Computer Vision. pp.
  6168--6177 (2021)

\bibitem{zheng2014patchmatch}
Zheng, E., Dunn, E., Jojic, V., Frahm, J.M.: Patchmatch based joint view
  selection and depthmap estimation. In: Proceedings of the IEEE Conference on
  Computer Vision and Pattern Recognition. pp. 1510--1517 (2014)

\bibitem{zhou2017scalemedian}
Zhou, T., Brown, M., Snavely, N., Lowe, D.G.: Unsupervised learning of depth
  and ego-motion from video. In: Proceedings of the IEEE conference on computer
  vision and pattern recognition. pp. 1851--1858 (2017)

\end{thebibliography}

\end{document}